# Recognizing Chinese Judicial Named Entity using BiLSTM-CRF

**Pin Tang[1], Pinli Yang[2,4], Yuang Shi[1], Yi Zhou[3], Feng Lin[1] and Yan Wang[1,5]**

[1] School of Computer Science, Sichuan University, Chengdu, Sichuan, China
[2] School of electronic information, Sichuan University, Chengdu, Sichuan, China
[3] Law School, Sichuan University, Chengdu, Sichuan, China

[4] co-first author
[5] wangyanscu@hotmail.com

**Abstract**. Named entity recognition (NER) plays an essential role in natural language processing systems. Judicial NER is a fundamental component of judicial information retrieval, entity relation extraction, and knowledge map building. However, Chinese judicial NER remains to be more challenging due to the characteristics of Chinese and high accuracy requirements in the judicial filed. Thus, in this paper, we propose a deep learning-based method named BiLSTM-CRF which consists of bi-directional long short-term memory (BiLSTM) and conditional random fields (CRF). For further accuracy promotion, we propose to use Adaptive moment estimation (Adam) for optimization of the model. To validate our method, we perform experiments on judgment documents including commutation, parole and temporary service outside prison, which is acquired from *China Judgments Online*. Experimental results achieve the accuracy of 0.876, recall of 0.856 and F1 score of 0.855, which suggests the superiority of the proposed BiLSTM-CRF with Adam optimizer.

## 1. Introduction

With the development of the judicial online systems, a large amount of judicial data of various kinds of cases and documents have been accumulated. How to give full play to the judicial data has become a hot topic of concern. Natural language processing (NLP) is a sensible way to deal with the data, of which named entity recognition (NER) is an indispensable subtask [1]. NER task is composed of two parts: detecting the entity boundaries and classifying the entities into predefined categories such as Name, Judicial Organization, and Location [2]. Chinese judicial NER can reduce the heavy burdens of related staff, improve the efficiency of the judicial industry, and help achieve information sharing.

However, different from other languages such as English, the Chinese language doesn't have spaces between words and lacks semantic features like uppercase, which makes Chinese NER more difficult [3]. Furthermore, judicial entities are mostly nested, making structures complicated, e.g."成都市中级人民检察院 (Chengdu Intermediate People's Procuratorate)" is supposed to be categorized into"Judicial Organization", but"成都市 (Chengdu)" may be categorized into"Location" in other text. Additionally, there are considerable terminologies, e.g."被告人(defendant)". All these factors conspire Chinese judicial NER towards a challenging task.

Only a small number of studies on Chinese NER are available, likely due to these challenges mentioned above. Earlier studies can be categorized into two groups: 1) dictionary- and rule-based methods [4]. Such methods require experts to manually create rule templates and use patterns and string matching to recognize named entities. 2) Statistic-based methods, including conditional random field [5], maximum entropy model [6], hidden Markov model [7] and support vector machine [8], etc. Both the two kinds of methods rely heavily on the corpus and require experts to manually extract features from the dataset [9]. However, the portability and accuracy of the two types of methods are relatively poor.

The development of deep learning, especially the proposed distributed word representation [10], sheds light on natural language processing [11-13]. And deep learning-based NER methods have also achieved some results [14]. However, the deep learning-based methods independently predict each character with the given features, without considering the labels that have been predicted previously, which may cause the predicted label sequence to be invalid and thus reduce the accuracy of entity recognition results [15].

In this paper, we propose to use BiLSTM-CRF model to recognize the Chinese judicial entity and utilize Adam optimizer for further accuracy improvement. BiLSTM-CRF not only retains the advantages of the deep learning-based methods i.e., using a character as the basic processing unit and assigning a label to each character, but also introduces some constrains to avoid the problem in the deep learning-based method mentioned above. Experiments are carried out extensively on documents acquired from *China Judgments Online*, and the experimental results indicate that our proposed method is satisfactory and promising.

The rest of the paper is organized as follows. Section 2 gives an illustration of the methodology and the architecture of our proposed BiLSTM-CRF. Then, experiments and experimental results are shown in Section 3. At last, we make a conclusion about this paper in Section 4.

## 2. Methodology

### 2.1. Bi-directional long short-term memory

Recurrent neural network (RNN) [16] is a typical deep learning model that can theoretically process series data and learn context information of any length. However, we can observe that if the length of the data is too long, the vanishing-gradient problem will occur and then the optimization cannot be continued. Thus, it can be concluded that the RNN is length-dependent.

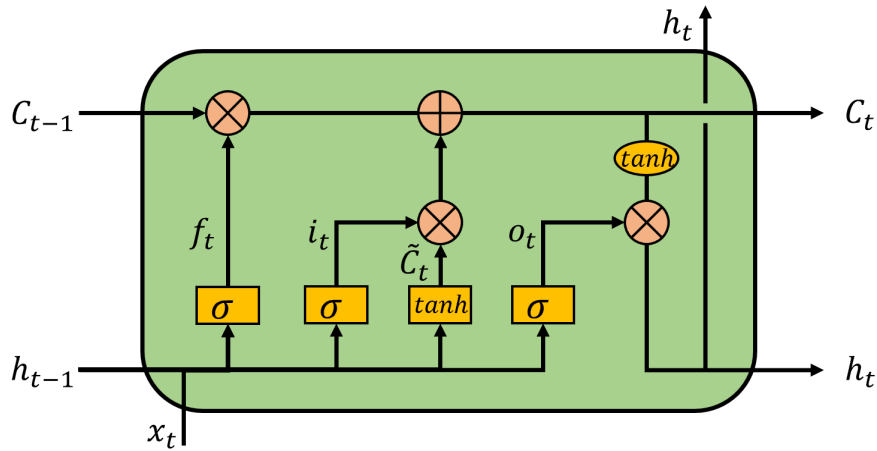

**Figure 1.** The structure of the LSTM cell [17].

Motivated by this shortcoming of RNN, long short-term memory (LSTM) was proposed in [17], which is an improved RNN model. LSTM can be regarded as a network of LSTM cells illustrated in Fig.1.

The internal calculation formula of the LSTM cell can be defined as follows:

$$\begin{aligned}
f_t &= \sigma(W_f[h_{t-1}, x_t] + b_f) \\
i_t &= \sigma(W_i[h_{t-1}, x_t] + b_i) \\
o_t &= \sigma(W_o[h_{t-1}, x_t] + b_o) \\
\widetilde{C}_t &= \tanh(W_C[h_{t-1}, x_t] + b_C) \\
C_t &= f_t \odot C_{t-1} + i_t \odot \widetilde{C}_t \\
h_t &= o_t \odot \tanh(C_t)
\end{aligned} \qquad (1)[17]$$

where $\sigma$ is Sigmoid function. $f_t$, $i_t$, $o_t$ respectively represent forget gate, input gate, and output gate. $W$ s denotes weight matrices and $b$ s are biases. $\widetilde{C}_t$ and $C_t$ are the candidate cell state and new cell state. $h$ means hidden state. $x_t$, $h_{t-1}$, $C_{t-1}$ are inputs, while $h_t$ and $C_t$ are outputs. Finally, we can obtain an output vector $(h_0, h_1, h_2, ..., h_{n-1})$ when there are $n$ LSTM cells.

It is obvious that the hidden state $h_t$ of the current LSTM cell relies on the previous hidden state $h_{t-1}$ but is irrelevant to the next hidden state $h_{t+1}$ i.e., information only flows in the forward direction in unidirectional LSTM. However, information from the backward direction is also useful to NER task. For instance, "成都市 (Chengdu)" in "成都市中级人民检察院 (Chengdu Intermediate People's Procuratorate)" may be mistakenly recognized as Location when using LSTM since the following "中级人民检察院 (Intermediate People's Procuratorate)" has no impact to the recognition of "成都市 (Chengdu)". On the contrary, Bi-directional LSTM (BiLSTM), which is composed of two LSTMs, i.e., forward LSTM and backward LSTM, can learn and merge information both from the forward and backward direction of a sentence. Hence BiLSTM understands the context better than LSTM. Assume that the output sequence of hidden states of the forward and backward LSTM respectively are $\overrightarrow{h_t}$ and $\overleftarrow{h_t}$, the context vector can be represented by catenating the two hidden vectors as $h_t = [\overrightarrow{h_t}; \overleftarrow{h_t}]$.

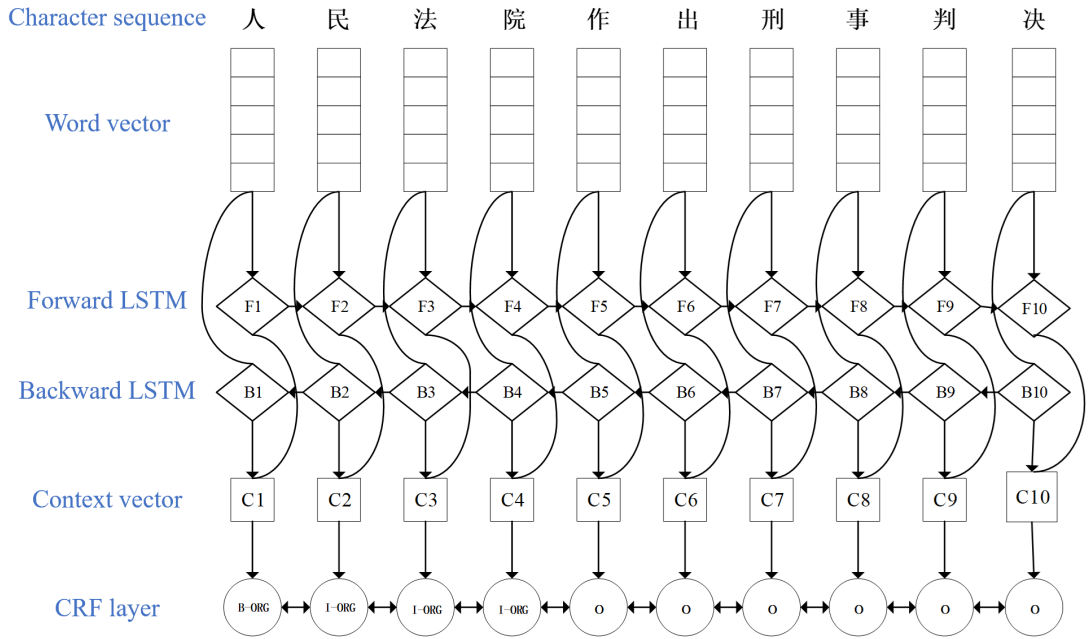

**Figure 2.** The architecture of BiLSTM-CRF.

*2.2. Conditional random field*

Adjacent labels usually have strong dependencies in the labeling task. For example, the I-ORG label should not follow the B-LOC label. With this issue in mind, we propose to utilize conditional random field (CRF) to jointly decode the labels for a given input sentence. CRF is a sequence marking model and was proposed in [18], which takes the advantages of the maximum entropy model and the hidden Markov model. It has superiority in sequence labeling tasks because it takes the order and correlations between labels into consideration. Hence, we use the CRF layer to get the optimal sequence of labels given the sequence of predictions $y = (y_1, y_2, y_3, ..., y_n)$ with the sentence $X$.

Let $A$ denotes the transition matrix in CRF layer, in which $A_{y_{i-1}, y_i}$ means the transition score from the label $y_{i-1}$ to the label $y_i$. And $P$ is the output score matrix of the BiLSTM. $P_{i,y_i}$ represents the confidence score of the word $i$ belonging to the label $y_i$. We can define score $s(X, y)$ as Equation 2.

$$s(X, y) = \sum_{i=1}^{n} P_{i,y_i} + \sum_{i=1}^{n} A_{y_{i-1}, y_i} \tag{2}$$

It is obvious that the higher $s(X, y)$ is, the prediction is more accurate. So, we can obtain the most reliable output using Equation 3.

$$y^* = \arg\max(\log(soft\max(s(X, y)))) \tag{3}$$

*2.3. BiLSTM-CRF*

The architecture of BiLSTM-CRF is illustrated in Fig 2. First, all the characters are converted into word vectors and are used as the input of BiLSTM. Then, $\overrightarrow{h_t}$ and $\overleftarrow{h_t}$ will be respectively calculated in forward and backward LSTM by Equation 1, and we can get the final hidden vector by catenating $\overrightarrow{h_t}$ and $\overleftarrow{h_t}$ as $h_t = [\overrightarrow{h_t}; \overleftarrow{h_t}]$. Thus, the hidden states of the whole sentence can be presented as $(h_0, h_1, h_2, ..., h_{n-1})$.

After that, we use the Equation 4 to map the hidden vector to a $k$-dimensional vector, where $k$ is the number of tags.

$$P_i = \tanh(W_i h_i + b_i) \tag{4}$$

where the $W_i$ and $b_i$ are trainable parameters. Hence, we can easily use $P_{i,y_i}$ to present the probability that the word $i$ belongs to the label $y_i$.

Next, we can combine the output of BiLSTM, i.e., the matrix $P$, with the parameters in CRF, i.e., the matrix $A$, through Equation 2 and obtain the score of the labels of the whole sentence. Finally, we utilize the Equation 3 in Section 2.3 to find the optimal labels of the sentence, i.e., $y^*$.

## 3. Experiments and results

*3.1. Dataset and data pre-processing*

In this paper, we conducted our experiments on 1000 judgment documents acquired from China Judgments Online, including commutation, parole and temporary service outside prison. Concretely, 600 judgment documents are selected as training set, 200 as validation set, 200 as test set. Especially, the format of all the documents are standardized by removing spaces and being annotated using BIO labels by corpus annotation tool YDEEA. As shown in Table 1, we defined 5 kinds of named entity i.e., Name, Location, Judicial Organization, Docket Number and Crime Type, as well as 11 kinds of label.

**Table 1.** The categories of BIO label.

| Label | Details |
| --- | --- |
| B-PER | The beginning of Name |
| I-PER | The rest of Name |
| B-LOC | The beginning of Location |
| I-LOC | The rest of Location |
| B-ORG | The beginning of Judicial Organization |
| I-ORG | The rest of Judicial Organization |
| B-NUM | The beginning of Docket Number |
| I-NUM | The rest part of Docket Number |
| B-CRI | The beginning of Crime Type |
| I-CRI | The rest of Crime Type |
| O | Not judicial entity |

*3.2. Implementation details*

We implemented our network with TensorFlow. Model training and testing were conducted on an NVIDIA GeForce GTX 1080 Ti. To be specific, the dimension of word vector is empirically set to 200, and the maximum sequence length is set to 300. The number of epochs is 300 and the batch size is 16. Dropout and the initial learning rate of the Adam optimizer are experimentally set to 0.5 and 0.001.

*3.3. Evaluation criteria*

As stated in Section 1, NER task involves two parts, entity boundary detection and entity type classification. Only if both boundary and type are identical to the ground truth, then the task can be considered correctly finished. We adopt four metrics, including precision, recall and F1 score, which is widely utilized in NER and can be expressed as Equation 4-6.

$$precision = \frac{TP}{TP + FP} \quad (5)$$

$$recall = \frac{TP}{TP + FN} \quad (6)$$

$$F1 = \frac{2 \times precision \times recall}{precison + recall} \quad (7)$$

where true positive (TP) denotes the number of judicial named entities which are recognized and match the ground truth, while false positive (FP) means the number of characters which are recognized as named entity but do not match the ground truth, and false negative (FN) represents the number of characters that are annotated as ground truth but are not recognized but our model. It is obvious that the higher precision, recall and F1 score, the better the model performs.

*3.4. Experiment results*

*3.4.1. Evaluation of different optimizers*

During the training process, every optimizer possesses its own advantages and disadvantages and is likely to influence the performance of the proposed network, so it is of great value to find out which optimizer fits in our model best. We therefore conducted a series of experiments using BiLSTM-CRF with the three most prevailing optimizers: Adam, GD and RMSprop. The experimental result in terms of precision, recall and F1, is summarized in Table 2.

**Table 2.** Comparison among different optimizers.

| Optimizer | Precision | Recall | F1 |
|---|---|---|---|
| **Adam** | **0.876** | **0.858** | **0.855** |
| GD | 0.745 | 0.646 | 0.688 |
| RMSprop | 0.839 | 0.794 | 0.813 |

We can observe that all the value of the evaluation metrics is higher when using Adam rather than GD and RMSprop, which is a manifestation of the superiority of Adam. Thus, we choose Adam as the optimizer in the training process.

*3.4.2. Result Analysis*
Table 3 gives an illustration of the performance of our model on the five types of entities. It is obvious that the model achieves highest precision, recall and F1 score when evaluated by Location, while lowest when evaluating on Docket Number. This may be due to the complicated structure of the Docket Number, which contains numbers, characters and parentheses, making it difficult to find a general rule.

**Table 3.** Different types of named entity recognition results.

| Type of entity | Precision | Recall | F1 |
|---|---|---|---|
| Name | 0.890 | 0.869 | 0.860 |
| **Location** | **0.897** | **0.870** | **0.865** |
| Judicial Organization | 0.880 | 0.832 | 0.848 |
| Docket Number | 0.845 | 0.813 | 0.826 |
| Crime Type | 0.855 | 0.820 | 0.846 |

*3.4.3. Comparison between BiLSTM-CRF and GRU-CRF*
To further investigate the superiority of BiLSTM-CRF, we conduct a comparative experiment using BiLSTM-CRF and GRU-CRF. Gate recurrent unit (GRU) [19] is an improved variant of LSTM, only consisting of update gate and reset gate. The experimental result is shown in Table 4. Note that, the precision, recall and F1 score of our method are respectively increased by 0.038, 0.06 and 0.028 when compared with GRU-CRF, which demonstrates our BiLSTM-CRF achieving better entity recognition than GRU-CRF.

**Table 4.** Comparison between among different optimizers.

| Model | Precision | Recall | F1 |
|---|---|---|---|
| GRU-CRF | 0.838 | 0.798 | 0.827 |
| **BiLSTM-CRF** | **0.876** | **0.858** | **0.855** |

## 4. Conclusion

Chinese NER remains to be challenging since the inherent characteristic of Chinese, especially in judicial filed which requires high accuracy. In this work, we propose to use a novel neural network for Chinese judicial NER, which incorporates BiLSTM and CRF. As an improved variant of RNN, BiLSTM can alleviate the length-dependent problem. However, BiLSTM do not consider the strong constrains among labels. Hence, we additionally introduce CRF to obtain the optimal labels. For further performance promotion, Adam is chosen as the optimizer during training process. The effectiveness of BiLSTM-CRF is extensively verified by conducting comparison experiments. And this model can be beneficial for judicial information retrieval, entity relation extraction, and knowledge map building. However, there are still some possibilities remain to be explored. For example, the dataset we used for model training is limited. In the near future, more reliable dataset will be acquired and more effort will be invested to improve the entity recognition accuracy and we will try to apply our method to other domains, such as clinical field.